\documentclass[letterpaper, 10 pt, conference]{ieeeconf}  

\usepackage{microtype}
\usepackage{graphicx}
\usepackage{subfigure}
\usepackage{mathtools}
\usepackage{hyperref}

\usepackage{tabularx}
\usepackage{multirow}

\usepackage{booktabs} 
\usepackage{amsfonts} 
\usepackage{nicefrac} 
\usepackage{microtype} 
\usepackage{xcolor}
\usepackage{xspace}
\usepackage{algorithm}
\usepackage{algorithmic}

\usepackage{comment}
\usepackage{bm}
\usepackage{bbm}
\usepackage{breqn}
\usepackage{authblk}
\usepackage{xspace} 
\usepackage{pifont}
\newcommand{\greencheck}{{\ding{52}}}
\newcommand{\redcheck}{{\ding{54}}}
\usepackage[capitalize,noabbrev]{cleveref}

\IEEEoverridecommandlockouts                              

\usepackage{amsmath} 
\usepackage{amssymb}  

\newcommand{\modelname}{{\textit{TrustNavGPT}}\xspace}

\title{\LARGE \bf
\textit{TrustNavGPT: }Modeling Uncertainty to Improve Trustworthiness of Audio-Guided LLM-Based Robot Navigation
}

\author{Xingpeng Sun$^{1}$, Yiran Zhang$^{1}$, Xindi Tang$^{1}$, Amrit Singh Bedi$^{2}$, Aniket Bera$^{1}$
\thanks{$^{1}$
        Purdue University, West Lafayette, IN 47907, USA
        {\tt\small {sun1223, zhan5058, tang666, aniketbera}@purdue.edu}}%

\thanks{$^{2}$
        University of Central Florida, Orlando, FL 32816, USA
        {\tt\small {amritbedi}@ucf.edu}}%
}

\begin{document}

\maketitle
\thispagestyle{empty}
\pagestyle{empty}

\begin{abstract}

Large language models (LLMs) exhibit a wide range of promising capabilities -- from step-by-step planning to commonsense reasoning --that provide utility for robot navigation. However, as humans communicate with robots in the real world, ambiguity and uncertainty may be embedded inside spoken instructions.  While LLMs are proficient at processing text in human conversations, they often encounter difficulties with the nuances of verbal instructions and, thus, remain prone to hallucinate trust in human command. 
In this work, we present \modelname, an LLM-based audio-guided navigation agent that uses affective cues in spoken communication—elements such as tone and inflection that convey meaning beyond words—allowing it to assess the trustworthiness of human commands and make effective, safe decisions. Experiments across a variety of simulation and real-world setups show a 70.46\% success rate in catching command uncertainty and an 80\% success rate in finding the target, 48.30\%, and 55\% outperform existing LLM-based navigation methods, respectively. Additionally, \modelname shows remarkable resilience against adversarial attacks, highlighted by a 22\%+ less decrease ratio than the existing LLM navigation method in success rate. Our approach provides a lightweight yet effective approach that extends existing LLMs to model audio vocal features embedded in the voice command and model uncertainty for safe robotic navigation. For more information, visit the \href{https://xingpengsun0.github.io/trustnav/}{TrustNav project page}.


\end{abstract}

\section{INTRODUCTION}


Recent advances in Large Language Models (LLMs), such as GPT-4 \cite{achiam2023gpt} or Gemini \cite{team2023gemini}, and Robotics have shown significant improvement for human-robot interactions (HRI) areas such as task planning \cite{liu2023llm+, li2023interactive, ren2023robots}, or social navigation \cite{shah2023navigation, shah2022lmnav,zhou2023navgpt}. It is crucial for robots to emulate how humans interact and form opinions about each other, including assessments of credibility and trust, and understand human uncertainty to ensure safe and efficient actions \cite{lewis2018role, schaefer2016measuring}. 
When humans interact, they subconsciously form opinions about one another, including judgments about vocal credibility and trust. These perceptions impact their decision-making in collaborative settings. For instance, imagine a scenario in which two individuals, both unfamiliar with a theme park, are interacting. One asks for directions to an attraction entrance, and the responder, uncertain of the way, provides unclear instructions. The inquirer, drawing on extensive experiential knowledge, can discern the uncertainty not only from the words but also from the hesitant vocal nuances. Consequently, they choose not to rely solely on this dubious guidance. In contrast, a robot lacking this nuanced reasoning capability would follow the instructions without question, potentially resulting in failure to reach the intended destination. 

To model human uncertainty, KnowNo \cite{ren2023robots} proposes an LLM-based planner and asks humans for clarification when needed, but it is only built on analyzing semantic uncertainty. However, in intricate settings such as theme parks, due to unfamiliarity with the space and spatial anxiety \cite{chan2012objects}, humans' guidance can be vague or uncertain, affected not just by the choice of words but also by the subtleties in their voice \cite{prestopnik2000relations, golledge2003human}. Current LLMs \cite{achiam2023gpt, team2023gemini} provide capabilities for converting speech to text, but this process often omits important vocal characteristics, leading to a significant loss of information that could indicate uncertainty. This gap in capturing vocal nuances limits the LLMs' capacity to accurately judge the reliability of voice-based commands and successfully navigate to the target, underscoring the necessity for advancements that can interpret and leverage these vocal cues in the realm of human-robot cooperation.

\begin{figure}[t]
\centering
\centerline{\includegraphics[width=\columnwidth]{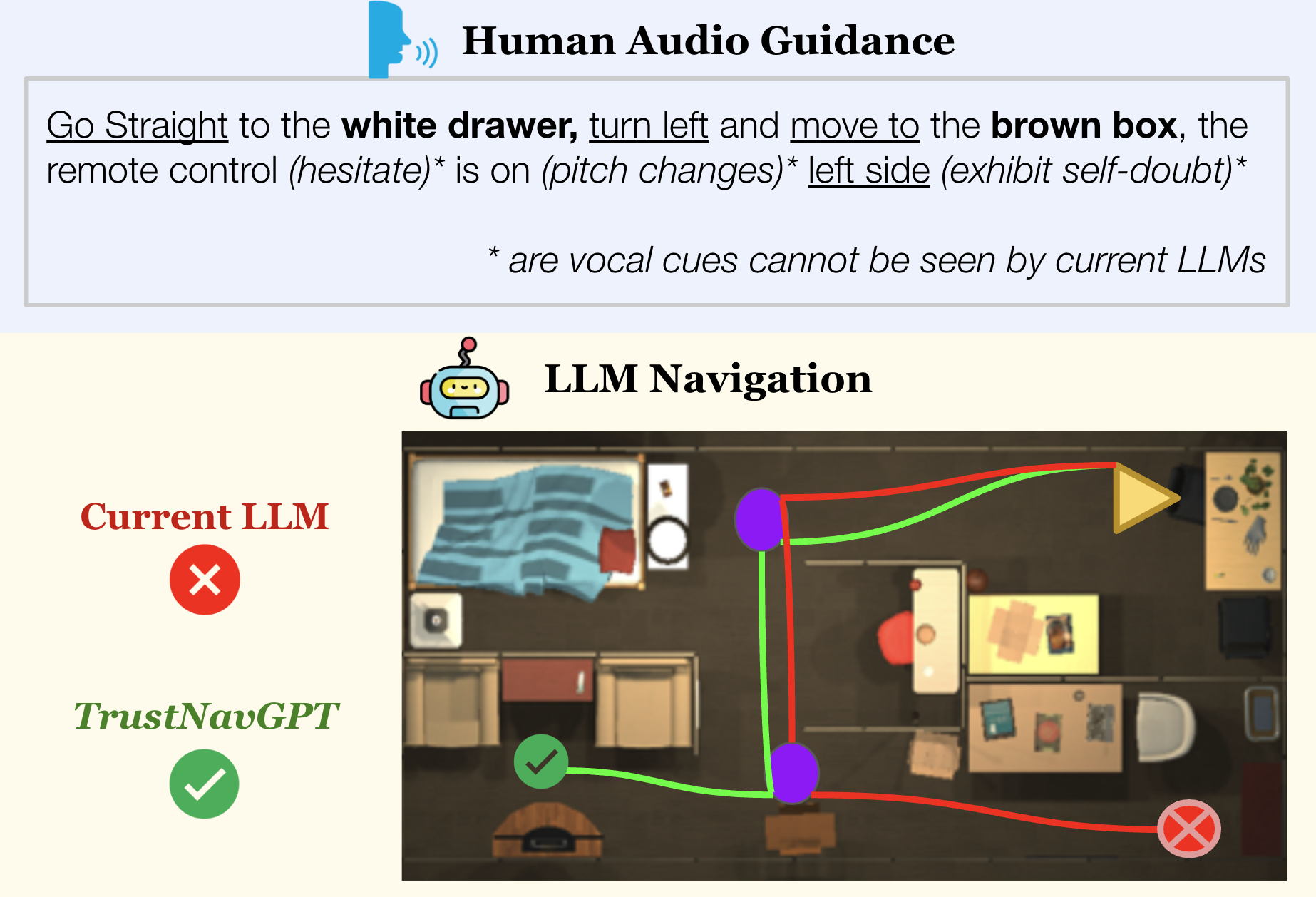}}
\caption{\textit{The current navigation methods using Large Language Models (LLMs) struggle with making accurate decisions when faced with ambiguous audio instructions. Our strategy involves affective cues from spoken communication into LLMs, enabling them to evaluate the reliability of human instructions from the semantic and vocal uncertainty, thus allowing for safe and successful navigation.}}
\label{fig:cover}
\vskip -0.2in
\end{figure}

\begin{figure*}[t]
\centering
\centerline{\includegraphics[width=\textwidth]{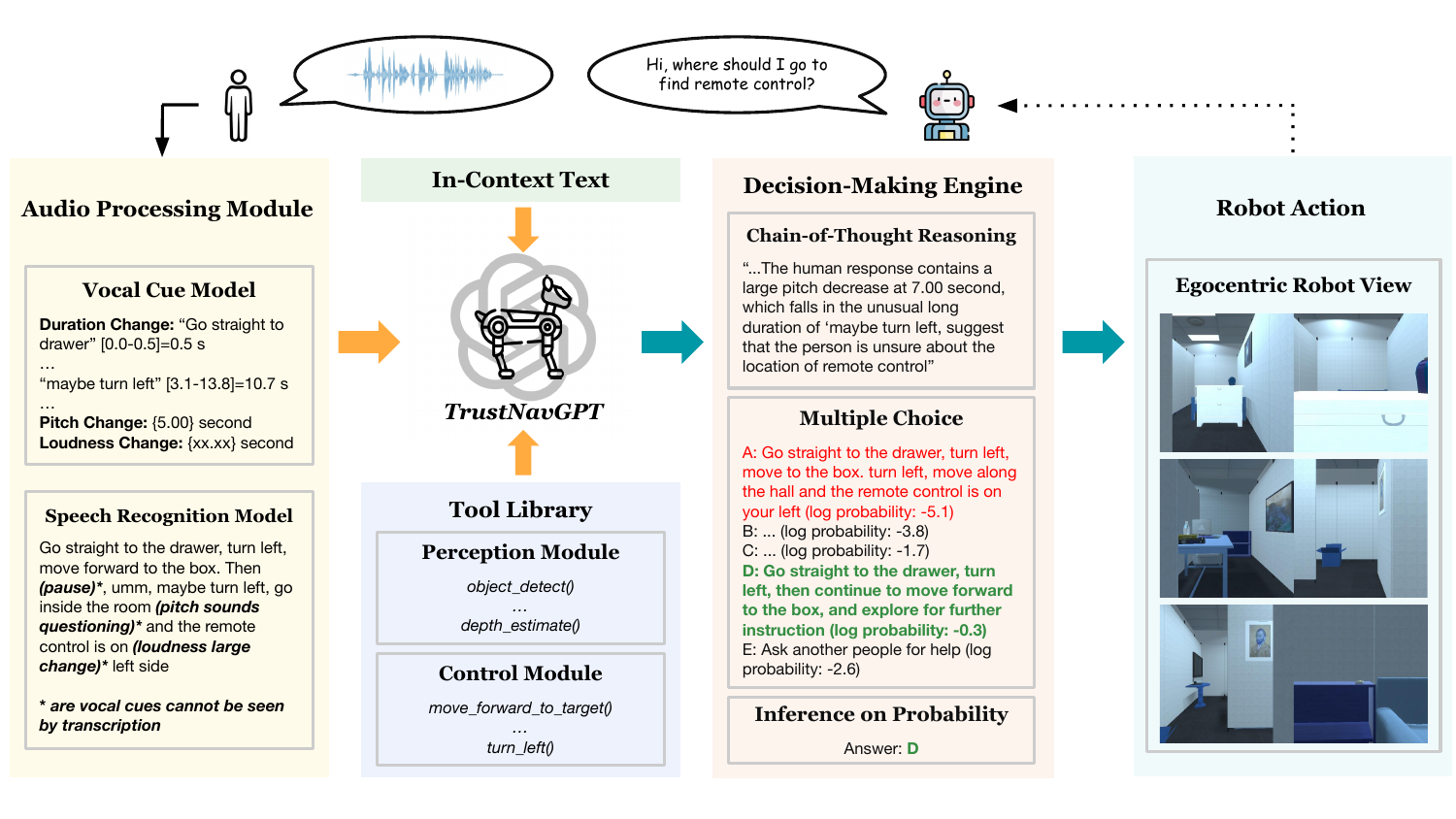}}
\caption{Overview: Human audio goes through an audio-processing module that transcribes it, while a vocal cue model identifies three essential affective cues. We then prompt a language model to generate five possible next-step actions, selecting the choice based on the next token logit probability. Notably, semantic transcription alone leads to the red choice, but incorporating the vocal cue results in the green choice being selected. Finally, a tool library translates the chosen language instruction into agent actions for navigation.}
\label{fig:overview}
\vskip -0.2in
\end{figure*}

Taking a step towards more human-like social navigation, we propose \modelname, a cognitive agent empowered by LLMs. The fundamental insight of our approach lies in the integration of both audio transcription and affective vocal features, including pitch, loudness, and speech rate, to improve robot ability in audio-guided navigation under uncertainty. Moreover, as LLM is good at high-level task planning but not good at low-level control and motion planning, we also propose a tool library that gives the LLM decision-making engine the ability to control the robot based on planning and visual perception. When the confidence of human guidance is reasoned as low, the robot undertakes scene exploration by categorizing objects in each direction and formulating conjectures based on each group (e.g. when the robot is tasked to find the microwave and sees a dishwasher on the left, it will infer left as kitchen and explore left side for target, instead of following human's uncertain command to ``turn right''). These components, coordinated by LLMs, create an automatic robotic system that navigates human audio uncertainty.
Our main contributions are summarized as follows:

\begin{enumerate}
\item  Our work introduces a layer of interpretation by examining not just the content of human speech but also the manner in which it is conveyed. The \modelname approach significantly refines LLMs' proficiency in interpreting human uncertainty within navigational contexts, evidenced by achieving over an 80\% success rate in robot navigation tasks. This integration allows for a more nuanced synthesis of information, paving the way for intelligent conversational systems, to better understand and act upon ambiguous human instructions.  

\item Integrate a motion planning tool library that translates high-level LLM language commands into robot actions, dynamic perception, and prediction, which can be accessed through function calls, to facilitate a human-like, audio-guided navigational capability in robots.


\item We conduct experiments on a large-scale Disfluent Navigational Instruction Audio Dataset \cite{sun2024beyond}, RoboTHOR simulation environment \cite{RoboTHOR}, and also real-world setup, to show that \modelname significantly surpasses existing LLM-based navigation techniques, by a 55\% improvement in achieving successful target arrival under conditions of human navigational uncertainty with 70\%+ closer to the target, indicating a substantial enhancement in navigational efficiency and precision. Detailed ablation studies on heterogeneous parts of our architecture are also provided, pointing to areas for future works.
\end{enumerate}

\section{Related Work}

\subsection{Large Language Model for Robotic Navigation}
With remarkable proficiency in commonsense reasoning and planning, Large Language Models (LLMs) have been utilized for navigation-related contexts. Recent scholarly work has explored the integration of LLMs with visual inputs to map landmarks and subgoals mentioned in navigational commands \cite{shah2022lmnav, yu2023l3mvn}, the application of LLMs in facilitating sequential decision-making for zero-shot robot navigation \cite{dorbala2023can, zhou2023navgpt}, and also the investigation of LLMs for the semantic prediction of object locations, thereby enhancing navigational efficiency \cite{shah2023navigation}. Despite these advancements, including NavGPT \cite{zhou2023navgpt}, the current body of research predominantly considers only textual instruction and assumes the reliability of human input commands, overlooking scenarios where such instructions might be ambiguous or incorrect. Our study distinctively addresses this gap by evaluating human uncertainty through the analysis of both textual and vocal emotions in audio-based navigation instructions. 

\subsection{Large Language Model Agent}
Inspired by strong emergent capabilities of LLMs, such as zero-shot prompting and complex reasoning, LLM agent, a system with complex reasoning capabilities, planning skills, and the means to execute tasks, becomes popular \cite{wang2023survey}. Voyager \cite{wang2023voyager} is an LLM-embodied gaming agent that plays Minecraft without human intervention through lifelong learning. Agent Driver \cite{mao2023language} and Inner Monologue \cite{huang2022inner} integrate LLM into autonomous driving systems and robot planning by incorporating environment feedback and making the LLM able to execute action through a versatile function library. However, to the best of our knowledge, current LLM agent works do not take into account the affective emotion of human command, especially extracting vocal uncertainty from speech in audio-guided navigation scenarios.

\subsection{Uncertainty Quantification for Large Language Model}
A growing body of research investigated quantifying uncertainty due to LLM's hallucinations \cite{huang2023survey, xiao2019quantifying}. Entropy has been introduced as a method to model uncertainty in the large language model \cite{kuhn2023semantic, lin2023generating}, while conformal prediction \cite{quach2023conformal} is another method applied to quantify uncertainty for next-token prediction in Multiple Choice Question Answering(MCQA) setups \cite{kumar2023conformal, ren2023robots}. In our approach, we take advantage of these works and define a confidence score $\mathcal{C}(\rho)$ inspired by entropy, which builds on the MCQA setup and shows effectiveness in gauging the LLM confidence.

\subsection{Affective Analysis in Social Robotics}
For social robots to effectively coexist and interact with humans, it is imperative that they comprehend human emotional states for decision-making processes. Emotion understanding from speech for human-robot interaction has been studied in \cite{lakomkin2018robustness}, while deep reinforcement learning methods \cite{dorbala2021can} and cognition model \cite{eppe2016exploiting} has been used to understand textual ambiguities in natural languages. While, for speech, not only does textual transcription offer insights, but vocal cues also hold substantial information that can reveal human emotions and ambiguities. However, there has been limited research on the role of vocal nuances in audio-guided social navigation. \modelname proposes an LLM agent that analyzes both textual and vocal affective cues embedded within human audio commands, creating a more human-like robot system for social navigation.

\section{Methodology}

\subsection{Problem Formulation} 
In this section, we mathematically formalize the navigation problem towards a designated target location $\tau$ under uncertainties within navigational instruction, utilizing Large Language Models (LLMs). In the depicted scenario (Figure\ref{fig:overview}), a robot seeks navigational commands from a human, communicated through auditory means. This framework is adapted by seminal works in robot social navigation \cite{eppe2016exploiting, hu2019safe, lakomkin2018speech, francis2023principles}. Upon the robot's inquiry, a human articulates an auditory instruction $v \in \mathcal{V}$, where $\mathcal{V}$ represents the ensemble of auditory commands. This vocal input is transcribed into a textual format $\mathcal{W}$ through a pre-trained transcription model $\mathcal{T}: \mathcal{V} \rightarrow \mathcal{W}$, with $\mathcal{W}$ embodying the set of all feasible textual instructions. Simultaneously, $v$ is mapped to an affective cue set $\mathcal{K}$ via an affective cue model $\mathcal{AC}: \mathcal{V} \rightarrow \mathcal{K}$. The combination of textual and affective cues is denoted as:
\begin{equation}
    \mathcal{P}(\mathcal{V}) = \mathcal{W} \oplus \mathcal{K},
\end{equation}
which constitutes the prompt for the LLM. The LLM (denoted as $F_{LLM}$) thus elucidates a response planning sequence $S$ conditioned on chain-of-thought reasoning $D$:
\begin{equation}\label{fuc:response_seq}
    S = \{s_1, s_2, \ldots, s_k\}=F_{LLM}(\mathcal{P}(\mathcal{V}) | D), 
\end{equation}
where each intermediate action step $s_i$ is generated sequentially. We define the joint probability distribution of generating the sequence $S$ from $\mathcal{P}(\mathcal{V})$ as:
\begin{equation}
    P_{\Theta}(S|\mathcal{P}(\mathcal{V})) = \prod_{i=1}^{k} P_{\Theta}(s_i|s_1, s_2, \ldots, s_{i-1}),
\end{equation}
where $\Theta$ denotes the parameter set of the LM. In the case at time $k$ the response $s_k$ is ambiguous, the robot leverages help from its decision-making engine (details in section\ref{sec:engine}) based on the visual exploration of the current state environment $M$ and thus inference target location from the surrounding objects. The high-level planning sequence is translated into low-level executable commands $\mathcal{A}=\{\alpha^1, \alpha^2, \ldots, \alpha^k\}$, where $\alpha^i = \varphi(s_i)$ based on a tool library $\varphi$ (details in section\ref{sec:tool}). The objective is to successfully navigate to target $\tau$, represented by:
\begin{equation}
\left\{
    \begin{aligned}
        & \max P(S', \tau, M) \\
        & \min \left\lVert R(\mathcal{A}, M) - \tau \right\rVert,
    \end{aligned}
\right.
\end{equation}

\noindent to maximize success rate for a robot $R$ to arrive at target $\tau$ in environment $M$ utilizing a LLM-reasoned step sequence $S'$ with minimum distance to target $\tau$ after applying the execution sequence $\mathcal{A}$. Detailed overview is shown in Fig.\ref{fig:overview}.

\begin{figure}[thpb]
\centering
\centerline{\includegraphics[width=\columnwidth]{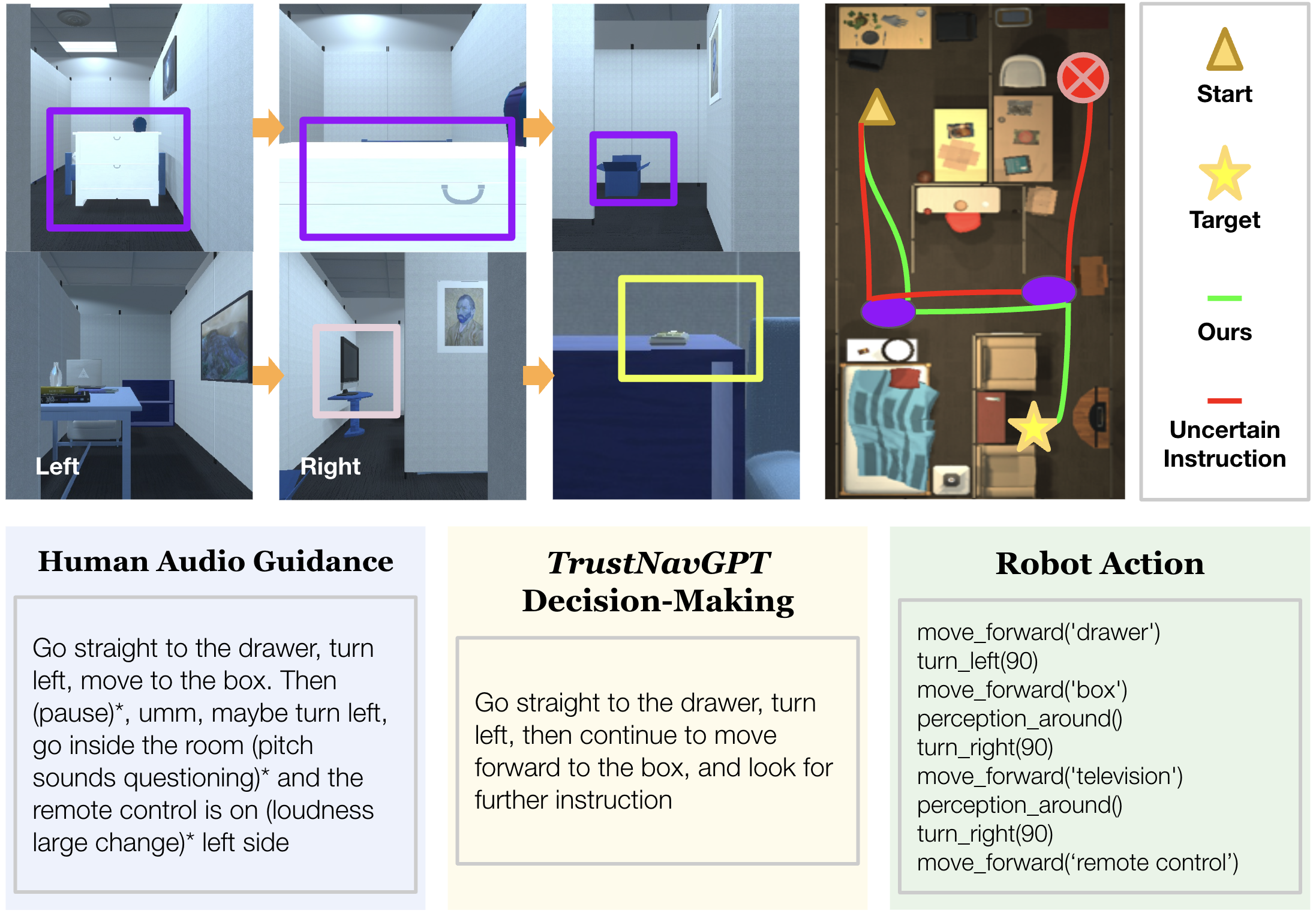}}
\caption{Illustration of action sequences. The purple box shows the reference object. At the point the human is ambiguous, the robot sees a television on the right-hand side(pick box), and thus reasons that the television is near to the remote control, then moves to the right side instead of following the human instruction. Notably, without uncertainty analysis, the LLM navigation path is shown in red, leading in the wrong direction. The navigation result of our method is shown in green, arriving at the target(yellow box) successfully.}
\label{fig:sim_step}
\vskip -0.2in
\end{figure}

\subsection{Audio Command Uncertainty Modeling} \label{sec:tvua}
In this section, we introduce the process of quantifying semantic and vocal uncertainty embedded in audio input. 

\textbf{Semantic Uncertainty:} With a human navigational guidance audio, we convert it into text using the open-source Whisper model \cite{radford2023robust}. Our analysis focuses on speech disfluencies as indicators of uncertainty, impacting the confidence measures of Large Language Models (LLMs). We identify three types of language uncertainty: \textit{\textbf{Ambiguous Word Choice}}, \textit{\textbf{Speech Repair}}, and \textit{\textbf{Hesitation Signs}}.

Ambiguous Word Choice encompasses phrases like ``probably," ``maybe," ``might," and ``I assume," indicating hesitancy or a lack of conviction \cite{dorbala2021can}. Speech Repair involves self-correction instances, such as ``Take a left turn, no no no, I mean take a right turn," reflecting errors in thought expression and potentially leading to confusion \cite{lou2019neural}. Hesitation Signs are pauses in speech, evident in expressions like ``Err, turn umm left," signaling uncertainty about forthcoming directions \cite{ogata2009use}. These patterns are prevalent in human communication when expressing doubt. We instruct LLMs to detect these cues, allowing for an adjustment in the confidence level regarding the reliability of instructions following these indicators.

\textbf{Vocal Uncertainty:} Beyond textual analysis, the prosody of spoken instructions—specifically \textit{\textbf{Pitch}}, \textit{\textbf{Loudness}}, and \textit{\textbf{Speech Rate}}—also serves as a marker of human uncertainty. Pitch is the rising intonation observed at the conclusion of phrases or sentences, resembling a questioning tone rather than an assertive statement \cite{prosodicsign,guyer2019speech}. Loudness is shown as fluctuations in volume levels, as it can suggest wavering of confidence \cite{prosodicsign,szekely2017synthesising}. In terms of speech rate, people might either slow down as they ponder their words or accelerate their speaking speed due to nervousness \cite{jiang2017sound,szekely2017synthesising}. Variations in speech rate serve as vocal markers reflecting changes in certainty levels \cite{guyer2019speech}. Our method measures the speech rate of each instruction segment within the audio and assesses whether notable duration has occurred. For instance, in a recording where each instruction phrase typically spans around one second, an elongated phrase extending for more than 6 seconds may signify hesitation and reduced confidence.

\begin{algorithm}
\caption{Vocal Uncertainty Modeling}
\label{alg:detection}
\begin{algorithmic}
   \REQUIRE audio $v$, loudness threshold $\theta_l$, pitch threshold $\theta_p$
   \ENSURE Timestamps of maximum loudness change ($t_l$) and pitch shift ($t_p$), speech rate of each instruction segment $r_s$
   \STATE $t_l \gets \text{argmax}_t \left\{ \delta_{l_t} \ \middle| \ (t, \delta_{l_t}) \in v, \forall \delta_{l_t}\geq \theta_l \right\}$ 
   \STATE $t_p \gets \text{argmax}_t \left\{ \delta_{p_t} \ \middle| \ (t, \delta_{p_t}) \in v, \forall \delta_{p_t}\geq \theta_p \right\}$   
   \STATE $r_s \gets \text{time} (s_i), \forall s_i \in S $
\end{algorithmic}
\end{algorithm}

As detailed in Algorithm\ref{alg:detection}, we detect the max change of loudness and pitch features $\delta_{l_t}$, $\delta_{p_t}$ in the audio clip. For the speech rate, we measure the time of each sub-instruction $s_i$. We then use force alignment, a technique that aligns text fragments with $\delta_{l_t}$, $\delta_{p_t}$, $r_s$, to synchronize the timestamps of vocal features with the literal instructions.

\subsection{Decision-Making Engine}\label{sec:engine}
The decision-making engine takes audio transcription and vocal analysis information as inputs, performs MCQA task planning, and scene direction conjecture, and eventually generates execution commands to navigate the robot to the target based on the potential ambiguous audio guidance.

\textbf{MCQA task planning: }To generate possible next steps based on $\mathcal{P}(\mathcal{V})$, the prompt consists both textual and vocal cue information of human audio command, we use Chain-of-Thought reasoning \cite{wei2022chain}, a prompting mechanism aims to emulate the human reasoning process, together with In-Context Learning \cite{yousefi2023context} by providing only few-shot examples with no explicit training. The process can be formulated as:
\begin{equation}
    \mathcal{D} = F_{LLM} (\mathcal{P}(\mathcal{V}), E),
\end{equation}
For in-context examples $E$, we fine-grained three examples that cover all human interaction scenarios: 1) textual uncertainty, 2) vocal uncertainty, and 3) both textual and vocal uncertainty. We ask the LLM to first identify potential uncertainty signals and then assess how this notable uncertainty will influence the subsequent decision-making process.

In the answering phase, the model reason to suggest a spectrum of five potential actions follows the Equation \ref{fuc:response_seq} in MCQA settings. Particularly, the first option (A) is a direct paraphrase of the transcription, excluding any uncertainty, representing the robot's choice to unconditionally accept the human audio-guided instruction. Options (B), (C), and (D) reflect various actions acknowledging uncertainty, while the last option (E) is always ``ask another person nearby for direction'', providing a reliable fallback in the decision-making process. Then, we predict the next-token log probability, the most commonly-used pre-training objective for causal language models, for the set of options $Y = \{`A', `B', `C', `D', `E'\}$ to select the label with the highest probability as the optimal planning sequence $S$. 

\textbf{Visual Scene Direction Conjecture: }
Given optimal planning sequence $S = \{s_1, s_2, \ldots, s_k\}$, if the audio $v$ sounds uncertain, then $s_k$ is always an ambiguous action, like ``look for more information at this location to plan navigation to target'', due to learning in-context examples. We employ the semantic knowledge embedded in language models in a tailored manner, utilizing it not just as a heuristic for search \cite{shah2023navigation}, but also as a way to visual grounding the trust of human guidance. Our decision-making engine deduces $k$th action based on the visual exploration of the current state environment $M$ denoted as $\Lambda (s_k,M)$. Specifically, the robot will segment $M$ into left, right, front, and three directions and categorize objects in each direction to hypotheses about their locations relative to the target. As illustrated in Figure \ref{fig:sim_step}, when the target is a ``Remote Control'' and the robot, upon detecting a ``Television'' to the right at a point of ambiguous human instruction, it deduces that the ``Television'' is likely near the ``Remote Control''. This inference leads it to prioritize its own decision-making directions $s_k'$ over less reliable human directions $s_k$. On the other hand, if a limited number of objects can be detected in $M$, the decision-making engine will resort to any other supervisor agent, such as asking help from a human, denoted as $\Gamma$, to yield a clarified action $S'=\{s_1, s_2, \ldots, s_k'\}$, where $s_k' = \Lambda (s_k,M)\oplus \Gamma(s_k)$. \modelname, therefore, leverages vision to better infer the target location, minimizing the attempt to ask humans and mimic a human-like social navigation process.

\subsection{Tool Library}\label{sec:tool}
While Large Language Models (LLMs) have demonstrated impressive proficiency in high-level task planning~\cite{wu2023tidybot,shah2023navigation}, bridging the divide from strategic planning to practical execution remains a significant hurdle. Recent studies~\cite{huang2023instruct2act} have explored the use of foundation models to directly generate executable action codes from linguistic instructions. However, this approach often encounters limitations in the speed of translating instructions to actions, and the resultant code accuracy is not assured. To address these issues and enhance the precision of action codes, we follow the idea of~\cite{wang2023voyager}, which creates a skill library to store and retrieve behaviors for gaming agents and introduces a customized robot navigation tool library. This library comprises a collection of functions specifically tailored to parse environmental data based on textual instructions and decompose complex language into robot-executable actions through dynamic function calls. This framework establishes a robust loop encompassing perception, planning, and action, thereby facilitating robots to execute tasks more reliably and efficiently. 

\textbf{Tool Functions.} We developed functions for dynamic perception and control to enhance our system's interaction with its environment. For the perception module, we implemented object detection and depth estimation using off-the-shelf pre-trained models, enabling the system to dynamically detect the object and provide information for robot action planning. For the control module, we employ few-shot learning techniques, enabling the language model to convert textual navigation instructions into pairs of actions and locations through function calls. This approach is operated through custom functions such as \texttt{move_forward_to_target()}, \texttt{turn_left()}, which guide the robot's movement through each step of the instruction sequence. Fig.\ref{fig:overview}\&\ref{fig:sim_step} illustrate this integrated process.

\section{Experiments and Results}
In this section, we demonstrate the effectiveness, few-shot learning, and characteristics of \modelname through extensive experiments on Disfluent Navigational Instruction Audio Dataset (DNIA)~\cite{sun2024beyond}, RoboTHOR simulation environment~\cite{RoboTHOR}, and real-world scenarios. First, we introduce the evaluation metrics and setup; then, we compare our methods with other LLM-based navigation methods in simulation environments. Finally, we conduct ablation studies to investigate the reasoning ability, perception ability, and effectiveness of each vocal cue in analyzing human audio uncertainty. Finally, we show robustness to LLM adversarial attacks. To be specific about the evaluation conditions, Tables \ref{table:wr_cs}, \ref{table:ablation_study}, and \ref{table:attack} present results under real-world setups, as our audio clips are recorded by humans in a quiet environment. Table \ref{table:simulation_result} presents results from the RoboTHOR simulation environment. 

\subsection{Uncertainty Detection}
To demonstrate our agent's capability in detecting human uncertainty, we utilize the DNIA dataset as detailed by \cite{sun2024beyond}. DNIA dataset \cite{sun2024beyond} encompasses a range of navigational disfluencies, comprising 500 audio clips divided into two categories: language uncertainty (LU) with 285 clips, and vocal tone uncertainty (VU) with 215 clips. LU clips feature semantic disfluencies, such as hesitations and language uncertainties, whereas VU clips include instances where the vocal tone, rather than the textual content, indicates uncertainty. For example, \textit{``Go straight to the drawer, turn left and move to the garbage can, the vase (hesitate) is on (pitch changes) your left''} is a VU command, and \textit{``Go straight to the drawer, turn left and move to the garbage can, the vase maybe is on your left''} is a LU command. Each clip is labeled with a user-study annotation (human annotation is a suggested method for LLM-based HRI evaluation \cite{bang2023multitask, chang2023survey}), which chooses the best choices from a set of five multiple-choice options designed to represent the uncertainty inherent in the audio. We calculate the \textbf{Prompt Selection Success Rate(PSSR)} and the \textbf{Confidence Score(CS)} for evaluation. We defined PSSR as:
\begin{equation}
    PSSR=\frac{p_{succ}}{p_{total}}
\end{equation}
where $p_{succ}$ denotes the number of instances where the LLM
chooses a correct next-step action that aligns with the user study annotation, and $p_{total}$ denotes the total number of audio clips that prompt the LLM.

Inspired from prior works\cite{kuhn2023semantic, lin2023generating} which use entropy as a method to quantify uncertainty in large language models, we define confidence score $\mathcal{C}(\rho)$ based on the model's $p_{\theta}$ probability distribution over the potential candidates $\{y^j\}_{j=1}^J$ against the ground-truth response distribution $\rho^*$ as a dirac-delta over the true response as $\rho^* = [0, 1 \cdots 0]$, assuming the second candidate to be the optimal $y^{j*}$ with $j*=2$ without loss of generality. This confidence score is inversely proportional to the Kullback–Leibler(KL) divergence between $\rho$ and $\rho^*$, as shown in the equation: 
\begin{align}\label{confidence_}
    \mathcal{C}(\rho)= \frac{1}{ \text{KL}(\rho, \rho^*)},
\end{align}
Notably, a higher confidence score—which indicates a lower KL divergence between $\rho$ and $\rho^*$—is desirable, as it signifies a closer alignment with the ground truth. Together with PSSR and CS, we show low bias and low variance in our uncertainty measurement.

\begin{table}[thpb]
\centering
\footnotesize
\caption{Uncertainty Measurement by Context and Audio Type}
\label{table:wr_cs}
\begin{tabular}{l|c|ccc}
\toprule
\multirow{2}{*}{\textbf{Method}}&\multirow{2}{*}{\textbf{Metric}}&\multicolumn{3}{c}{\textbf{Audio Category}} \\

\textbf{} & \textbf{} &\textbf{\textit{All}}& \textbf{\textit{VU}}& \textbf{\textit{LU}} \\
\midrule
\multirow{2}{*}{\textbf{Text-based LLM}}& \textbf{PSSR} &22.16\%& 22.79\%& 21.75\% 
\\
\textbf{} & \textbf{CS} & 0.7782 & 0.7656 & 0.7884
\\
\midrule
\multirow{2}{*}{\textbf{With CoT}}& \textbf{PSSR}& 49.30\%& 36.74\%& 58.60\%
\\
\textbf{} & \textbf{CS}& 0.9545 & 0.9553 & 0.9549
\\
\midrule
\multirow{2}{*}{\textbf{Ours}}& \textbf{PSSR}& \textbf{70.46\%}& \textbf{72.56\%}& \textbf{68.77\%}
\\
\textbf{} & \textbf{CS}& \textbf{1.1354} & \textbf{1.1189} & \textbf{1.1445}
\\
\bottomrule
\end{tabular}
\end{table}

\begin{table}[t]
\caption{Ablation Study for Various Vocal Cues}
\label{table:ablation_study}
\centering
\footnotesize
\begin{tabular}{ccc|ccc|l}
\toprule
\multicolumn{3}{c}{\textbf{Vocal Cue}}&\multicolumn{3}{c}{\textbf{PSSR}} & \textbf{CS} \\
\midrule
Pitch&Loudness&Speech Rate& \textbf{\textit{All}}& \textbf{\textit{VU}}& \textbf{\textit{LU}} &\textbf{All}\\
\midrule
\redcheck &\redcheck &\redcheck & 49.30& 36.74& 58.60 & 0.4965
\\
\greencheck &\redcheck &\redcheck&61.68&64.65&59.30 & 0.8322
\\
\redcheck &\greencheck &\redcheck& 63.07&61.40&64.21 & 0.9683
\\
\redcheck &\redcheck &\greencheck& 60.88& 61.40& 60.35 & 0.8467
\\
\greencheck &\greencheck &\redcheck& 67.47&70.23&65.26& \textbf{1.1282}
\\
\greencheck &\redcheck &\greencheck& 64.87& 66.51& 63.51 & 0.9076
\\
\redcheck &\greencheck &\greencheck& 65.47&67.91&63.51 & 1.0115
\\
\greencheck &\greencheck &\greencheck & \textbf{70.46}& \textbf{72.56}& \textbf{68.77} & 0.8227
\\
\bottomrule
\end{tabular}
\end{table}

Table \ref{table:wr_cs} demonstrates that our method exhibits reduced bias and reduced variance, outperforming both the single-modal transcription method and the reasoning-augmented approach. Integrating vocal affective cues, which allow LLMs to process how statements are spoken, markedly enhanced performance. The overall PSSR surged to 70.46\%, with a pronounced improvement in VU interpretation, evident in a 72.56\% PSSR. The overall confidence score surged to 1.135, improved by 45.8\% in comparison to existing LLM methods.

To elucidate the impact of individual components, we conduct a comprehensive ablation study on each category of vocal cues and report in Table \ref{table:ablation_study}. Each of the Pitch, loudness, and speech rate features can significantly enhance LLM's ability to discern vocal uncertainty, increasing the PSSR by 20\%+ and CS by 127\%. Regarding CS, the inclusion of vocal features consistently outperforms scenarios lacking these features and also surpasses those without reasoning. The combination of pitch and loudness features achieves the highest confidence score. This suggests that while the presence of multiple vocal features across different time frames may slightly diminish the model's confidence, it does not affect its accuracy in selecting the correct responses.

\begin{table*}[thpb]
\centering
\footnotesize
\caption{RoboTHOR Navigation Result \& Ablation Result on Different Module}
\label{table:simulation_result}
\begin{tabular}{l|cc|cc|cc|cc|cc}
\toprule
\multirow{2}{*}{\textbf{Method}}&\multicolumn{2}{c}{\textbf{SR}}&\multicolumn{2}{c}{\textbf{Steps}}&\multicolumn{2}{c}{\textbf{Path Distance}}&\multicolumn{2}{c}{\textbf{Distance to Target}}&\multicolumn{2}{c}{\textbf{SPL}} \\

\textbf{}&\textbf{\textit{LU}}&\textbf{\textit{VU}}&\textbf{\textit{LU}}&\textbf{\textit{VU}}&\textbf{\textit{LU}}&\textbf{\textit{VU}}&\textbf{\textit{LU}}&\textbf{\textit{VU}}&\textbf{\textit{LU}}&\textbf{\textit{VU}} \\
\midrule
Random Search &25\%&25\%&534.6&534.6&31.62&31.62&1.56&1.56&8.88\%&8.88\%
\\
LM-Nav~\cite{shah2022lmnav} &25\%&25\%&8.5&8.5&5.30&5.30&2.35&2.35&9.57\%&9.57\%
\\
\midrule\midrule
\textbf{\modelname w/o Vocal (Ours)}& 25\%&50\%&8.5&9.0&4.53&5.99&2.55&2.39&13.46\%&30.24\%
\\
\textbf{\modelname w/o Vision (Ours)}& 75\%&50\%&10.0&13.5&6.42&6.79&1.84&1.75&\textbf{33.30\%}&\textbf{36.32\%}
\\
\textbf{\modelname (Ours)}& \textbf{80\%}&\textbf{80\%}&\textbf{13.2}&\textbf{14}&\textbf{10.24}&\textbf{9.94}&\textbf{0.56}&\textbf{0.82}&32.05\%&35.47\%
\\
\bottomrule
\end{tabular}
\vskip -0.1in
\end{table*}

\subsection{Simulation Environment Robot Navigation}
To evaluate the effectiveness of \modelname in navigation, we adopt the LoCoBot and extensively test the audio-guided navigation performance in 10 different RoboTHOR indoor environments. For each test, we provide a piece of audio instruction with either semantic or vocal uncertainty that navigates the LoCoBot to a unique target instance in the environment. We evaluate 5 common robot navigation metrics: 

\noindent \underline{\textbf{Success Rate (SR)}}: 
The LoCoBot successfully found the target within its vision distance. The higher SR is better.

\noindent \underline{\textbf{Steps}}: 
the number of robot movement actions.

\noindent \underline{\textbf{Path Distance}}: 
the explore path length that takes the LoCoBot to find the target if it succeeds or execute navigation events if it fails.

\noindent \underline{\textbf{Distance to Target}}: 
The shortest path distance from the LoCoBot's final position to the target position. The smaller this metric is, the more successful the navigation method is.

\noindent \underline{\textbf{Success weighted by Path Length (SPL)}}:
\begin{align}\label{spl}
    SPL = \frac{1}{N} \sum_{i=1}^{N} \frac{S_i}{\max(p_i, l_i)},
\end{align}
SPL ranges from [0,1], where $N$ is the total number of evaluated tasks, $S_i \in \{0,1\}$ is the binary indicator of success, $l_i$ denotes the ground truth shortest path length, and $p_i$ denotes the actual path length of the agent in navigation. This metric indicates the efficiency of the actual path compared to the ground truth shortest path when the navigation task is successfully completed.

Note in Table\ref{table:simulation_result}, our method significantly outperforms the random method and existing LLM methods, achieving 80\% for SR, an overall 0.69-meter distance to target, and 33.76\% in terms of SPL. We also run a comprehensive ablation study on each section of \modelname architecture, as illustrated in the last three columns in Table\ref{table:simulation_result}. With no vocal cue analysis, just using few-shot in-context learning to teach LLM to detect semantic uncertainty results in a low 27.5\% success rate. The highest SPL of 34.81\% is achieved using \modelname without a perception module due to the fact that the perception toolbox will lead robots to explore the environment and navigate to the related objects based on inference, thus resulting in longer total path distance and then lower SPL. The highest success rate and nearest distance to the target are observed using a combination of the perception module and vocal cue module.

\begin{table}[thpb]
\centering
\footnotesize
\caption{Robustness to LLM Attack}
\label{table:attack}
\begin{tabular}{l|ccc}
\toprule
\multirow{1}{*}{\textbf{Method}}&\multicolumn{1}{c}{\textbf{PSSR}} & \textbf{Distance to Target} & \textbf{SPL} \\
\midrule
LM-Nav\cite{shah2022lmnav} + Few-Shot & 22.16\%& 2.47& 21.85\% 
\\
After Token Attack& 9.78\%& 2.72& 19.24\% 
\\
Decrease &\textbf{55.87\%}&\textbf{0.25}&\textbf{13.57\%}\\
\midrule
\modelname (Ours)&70.46\%&0.69&33.76\%
\\
After Token Attack& 46.90\%& 0.91& 31.39\%
\\
Decrease &\textbf{33.43\%}&\textbf{0.22}&\textbf{7.55\%}\\
\bottomrule
\end{tabular}
\vskip -0.15in
\end{table}
\subsection{Robustness toward Adversarial Language Model Attacks}
With the progression of language models, concerns like adversarial attacks and prompt manipulation have gained prominence \cite{Ribeiro2018SemanticallyEA, shayegani2023plug}. These attacks often involve simple token operations such as synonym replacement and misleading models into making errors \cite{morris2020textattack, li2020bert}.

We present the \modelname resistance to such adversarial tactics aimed at LLMs. Our attack involves initially paraphrasing a given transcript ($T1$) into a new form ($T2$), where uncertain terms are all swapped for more deterministic phrases. ($T2$) is then used to realign both vocal and textual prompts and replicate as one option for LLM to choose. This experiment underscores the current LLMs' dependency on textual semantics, overlooking the subtleties embedded in vocal expression. We compare the result between adding in-context examples of detecting textual uncertainty within navigation command to existing LLM navigation methods against our method and show the result in Table\ref{table:attack}. After the token attack, our approach exhibited a notably lower reduction of 33.43\% in PSSR, 0.22 in distance to target, and 7.55\% in SPL, all notably smaller than that of the existing LLM navigation method. This suggests that audio augmentation in our approach enables LLMs to resist text-based adversarial attacks and maintain safe capabilities for robot navigation.

\subsection{Real-World Exploration}
\begin{figure}[t]
\centering
\centerline{\includegraphics[width=\columnwidth]{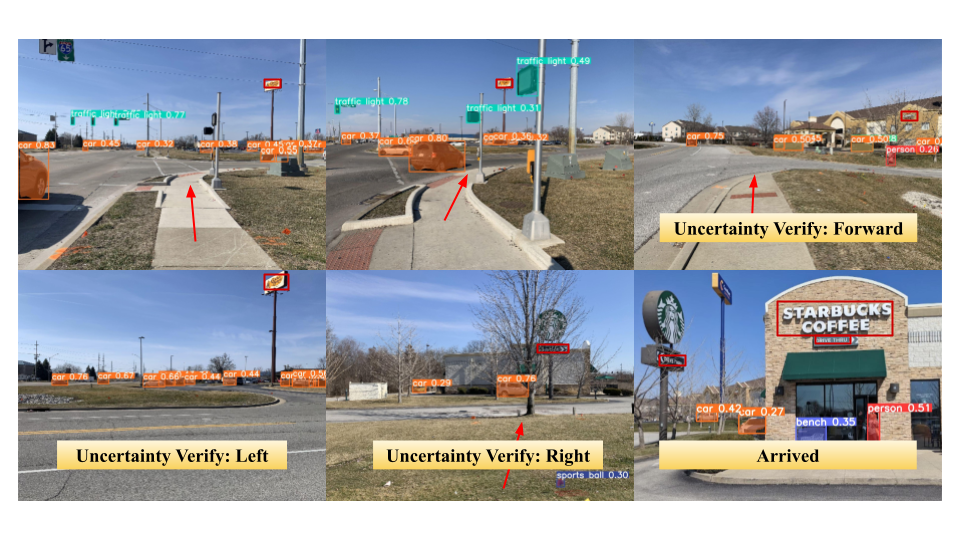}}
\caption{Real-world Navigation with vocal direction to Starbucks Coffee Shop. Successfully arrived at the target.}
\label{fig:realworld}
\vskip -0.2in
\end{figure}

\modelname underwent rigorous testing within real-world scenarios, employing YOLOv8 \cite{reis2023real} for object detection, the Tesseract Open Source Engine \cite{smith2007overview} for letter/word detection, and MiDas  \cite{Ranftl2022} for generating depth maps. At each time stamp, an image is captured and subsequently analyzed by YOLOv8 and Tesseract to ascertain the presence of target objects/words within the scene. If the target is not detected, the robot proceeds forward; otherwise, upon detection, the depth map for the identified target object is computed to determine if the robot should proceed forward or do the next turning task. If the object is close enough, a turning action is executed. A real-world demonstration illustrated in the accompanying Figure \ref{fig:realworld} shows how the robot does the mission of detected verbal instructions: ``walking straight until you see the traffic light; you wanna then turn left. Then, when you see a Comfort Suite, consider executing a uhhh...maybe a left turn, yea, you'll see a Starbucks drive-thru sign. Continue straight along this path; you'll see Starbucks coffee shop." Notably, uncertainty arose regarding the direction of the second left turn instruction, prompting the robot to analyze its surroundings by checking forward, left, and right perspectives to make sure it could arrive at the destination as expected. In this instance, the uncertain human instruction was rectified as the robot identified a drive-thru letter sign on its right-hand side, prompting a refined trajectory adjustment to turn right.

\section{CONCLUSION}
In our work, we present an LLM trust-driven audio-guided robot navigation agent \modelname, which effectively deals with potential uncertainty within human audio commands. Our findings highlight the improved planning, efficiency, and resilience achieved by integrating affective audio processing with large language models (LLMs) to improve navigation in social robots. As the integration of vocal and semantic analysis increases the computational overhead, which may
limit the deployment in low-resource settings or in real-time applications; and system's performance relies on the quality of audio input, Future works can include development of denoisy methods and more intelligent retrieval-augmented generation to improve the reliability and computation efficiency. We believe our work will encourage further exploration into aligning uncertainties with LLMs for the development of audio-directed robots.

\bibliographystyle{IEEEtran} 
\bibliography{ref}


\end{document}